\documentclass[letterpaper]{article} 
\usepackage{aaai25}  
\usepackage{times}  
\usepackage{helvet}  
\usepackage{courier}  
\usepackage[hyphens]{url}  
\usepackage{graphicx} 
\usepackage{natbib}  
\usepackage{caption} 
\usepackage{algorithm}
\usepackage{algorithmic}
\usepackage{amsmath} 
\usepackage{amssymb} 
\usepackage{booktabs} 
\usepackage{multirow} 
\usepackage{makecell} 
\usepackage{newfloat}
\usepackage{listings}
\DeclareCaptionStyle{ruled}{labelfont=normalfont,labelsep=colon,strut=off} 
\floatstyle{ruled}
\newfloat{listing}{tb}{lst}{}
\floatname{listing}{Listing}
\title{
An Efficient Framework for Enhancing Discriminative Models via Diffusion Techniques
}
\author {
    Chunxiao Li\textsuperscript{\rm 1}\equalcontrib,
    Xiaoxiao Wang\textsuperscript{\rm 2}\equalcontrib,
    Boming Miao\textsuperscript{\rm 1},
    Chuanlong Xie\textsuperscript{\rm 1},
    Zizhe Wang\textsuperscript{\rm 3},
    Yao Zhu\textsuperscript{\rm 3}\thanks{Corresponding author: Yao Zhu.}
}
\affiliations {
    \textsuperscript{\rm 1}Beijing Normal University, Beijing, China\\
    \textsuperscript{\rm 2}University of Chinese Academy of Sciences, Beijing, China\\
    \textsuperscript{\rm 3}Tsinghua University, Beijing, China\\
}

\usepackage{bibentry}

\begin{document}

\maketitle

\begin{abstract}
Image classification serves as the cornerstone of computer vision, traditionally achieved through discriminative models based on deep neural networks. Recent advancements have introduced classification methods derived from generative models, which offer the advantage of zero-shot classification. However, these methods suffer from two main drawbacks: high computational overhead and inferior performance compared to discriminative models. Inspired by the coordinated cognitive processes of rapid-slow pathway interactions in the human brain during visual signal recognition, we propose the Diffusion-Based Discriminative Model Enhancement Framework (DBMEF). This framework seamlessly integrates discriminative and generative models in a training-free manner, leveraging discriminative models for initial predictions and endowing deep neural networks with rethinking capabilities via diffusion models. Consequently, DBMEF can effectively enhance the classification accuracy and generalization capability of discriminative models in a plug-and-play manner. We have conducted extensive experiments across 17 prevalent deep model architectures with different training methods, including both CNN-based models such as ResNet and Transformer-based models like ViT, to demonstrate the effectiveness of the proposed DBMEF. Specifically, the framework yields a 1.51\% performance improvement for ResNet-50 on the ImageNet dataset and 3.02\% on the ImageNet-A dataset. In conclusion, our research introduces a novel paradigm for image classification, demonstrating stable improvements across different datasets and neural networks.
\begin{links}
    \link{Code}{https://github.com/ChunXiaostudy/DBMEF}
\end{links}
\end{abstract}

\section{Introduction}
\label{sec:intro}

\begin{figure}[t]
  \centering
  \includegraphics[width=1.0\columnwidth]{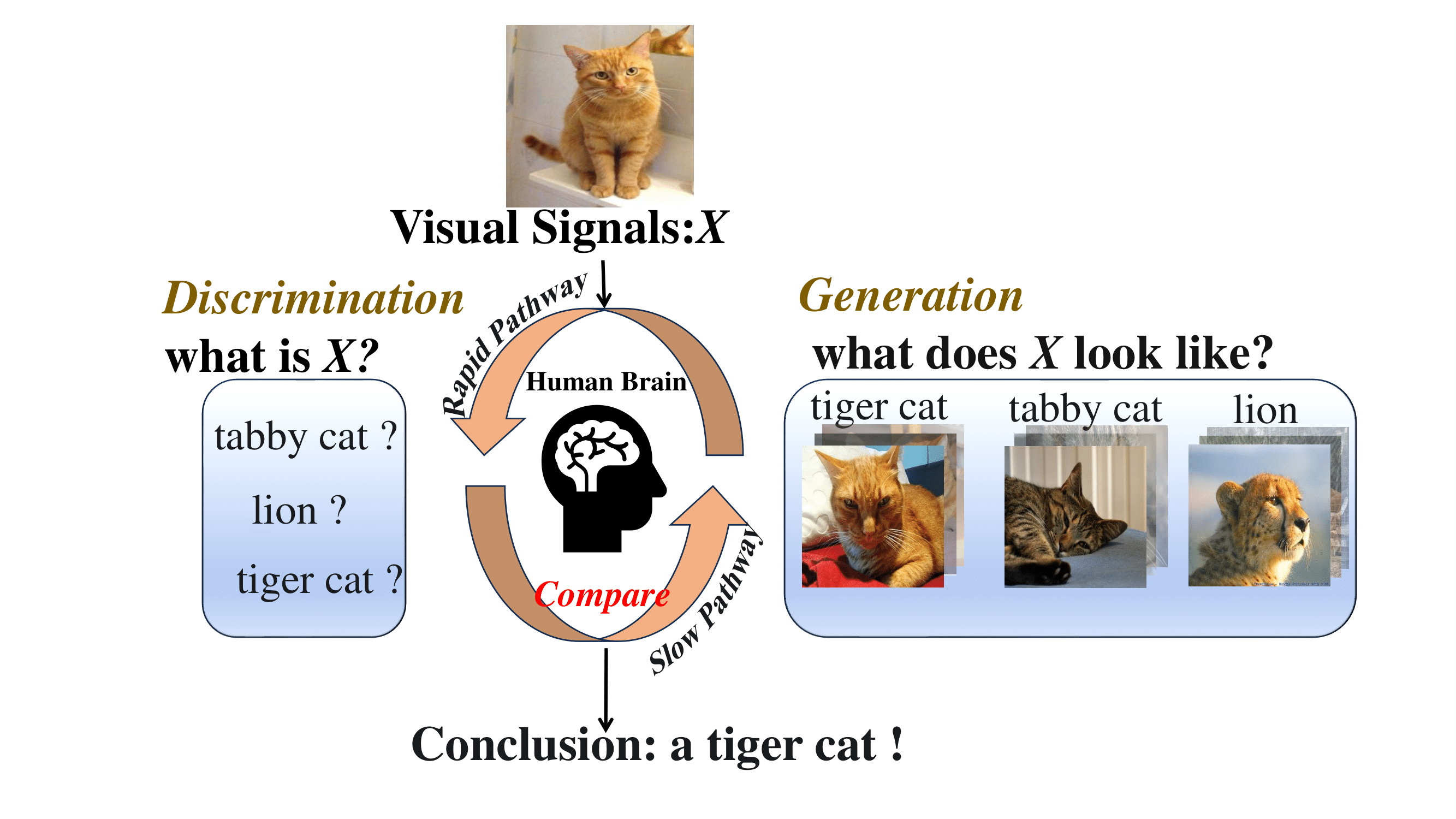}
  \caption{The process of the human brain handling visual signals is a dynamic interactive procedure. The rapid pathway transmits visual signals to the higher cortex to complete overall recognition, then assists the slow pathway in completing a ``guess-verify-guess-verify" cognitive linkage. The rapid pathway of the brain can be regarded as a discriminative process, proposing several possible guesses regarding the visual signals. The slow pathway's verification of these guesses can be approximately considered as the reclassification process of a generative model under given conditions.}
  \label{fig:image0}
\end{figure}

 Image classification stands as a foundational problem in computer vision, typically tackled through either discriminative model-based \cite{lecun1998gradient,he2016deep,touvron2021training} or generative model-based \cite{rish2001empirical,cheng2013comparing,han2022card} approaches: the former directly models the posterior probability $p(y|\boldsymbol{x})$ for image classification, while the latter first learns the joint distribution $p(\boldsymbol{x}, y)$ of the data and then leverages Bayes' theorem to convert it to $p(y|\boldsymbol{x})$.

In research on applying discriminative models for image classification, numerous meticulously designed network architectures have been proposed, such as VGG \cite{simonyan2014very}, ResNet \cite{he2016deep}, ViT \cite{dosovitskiy2020image}, DeiT \cite{touvron2021training}, and others. A well-trained ViT model can even achieve an astonishing accuracy of up to 88.59\% on ImageNet. However, this approach provides predictions in a single-step process, which means it lacks the capability to re-evaluate and refine uncertain predictions as humans do. This limitation potentially hinders further performance improvements for the model.

The task of image classification based on generative models is challenging because it requires modeling the conditional likelihood of each label for the images. Early work has explored the potential of generative models based on energy-based models \cite{zhao2020joint} or score-based models \cite{song2020score,zimmermann2021score,yoon2021adversarial} for image classification tasks; however, due to the greater demand for data and resources, such methods are primarily applied to smaller datasets like CIFAR-10, and there remains a significant gap in their widespread application. The recent emergence of large-scale text-to-image diffusion models \cite{ho2020denoising, rombach2022high} has greatly enhanced text-based image generation capabilities, inspiring researchers \cite{clark2024text,li2023your,chen2023robust} to use density estimation from diffusion models for zero-shot classification without additional training. To be specific, they perform conditional denoising across various categories on the test image, selecting the category that yields the best denoising outcome as the image's label. Although these methods can be applied to higher-resolution images, they still suffer from two drawbacks: their performance significantly lags behind discriminative models, and the application of diffusion models in classification is time-intensive. Even with accelerated sampling methods, classifying a single image from ImageNet of 512 $\times$ 512 pixels may take up to two hours to complete.

In response to the shortcomings of existing image classification approaches, we pose a question: Can we leverage the strengths of both discriminative and generative models to better accomplish the task of image recognition? This paper answers this question by drawing inspiration from the human brain's process of image recognition. As shown in Fig. \ref{fig:image0}, when the brain receives visual signals, the image information of objects is rapidly transmitted to the higher cortex through a rapid pathway, where a guess is made \cite{sillito2006always}. The result of the guess is then cross-verified with new inputs through feedback connections. The rapid pathway recognizes the object as a whole, and its results assist the slow pathway in recognizing local information of the object. Through such a repetitive process, the object is recognized. When faced with uncertain visual signals, the brain undergoes an alternation of information uploading and downloading, continuously engaging in a ``guess-verify-guess-verify" cycle\cite{sillito2006always,chen2014cell} until a definitive result is obtained.
Therefore, in this paper, inspired by the brain's signal recognition process that intricately combines the rapid and slow pathways for synergistic effect, we propose the Diffusion-Based Discriminative Model Enhancement Framework (DBMEF). This approach utilizes diffusion models to endow discriminative models with the ability to re-evaluate uncertain predictions. Specifically, within this framework, discriminative models first make a preliminary prediction on the test input, akin to the brain's rapid pathway. If the uncertainty of the prediction is low, it is considered the final output. However, if the prediction's uncertainty is high, the test sample is fed into the diffusion model. Through the generative model's powerful image understanding capabilities, the test sample undergoes a rethinking process, mirroring the brain's slow pathway. This methodology not only simulates the human brain's process of dealing with visual signals but also significantly enhances the classification accuracy of various discriminative models (e.g., ResNet, VGG, DeiT, ViT).

The contribution can be summarized as follows:
\begin{itemize}
 \item[$\bullet$] We propose a plug-and-play framework called Diffusion-Based Discriminative Model Enhancement Framework (DBMEF), which is inspired by  the coordinated cognitive processes of rapid-slow pathway interactions in the human brain during visual signal recognition.
 \item[$\bullet$] Experimental results have demonstrated that DBMEF can significantly enhance the classification accuracy and generalization capability of deep neural networks. The framework notably increases the performance of ResNet-50 by 1.51\% on the ImageNet dataset, and by 3.02\% on the ImageNet-A dataset.
 \item[$\bullet$] We provide an extensive and detailed ablation study on the proposed framework, uncovering that an overly intensified negative control factor $\lambda$ adversely impacts the efficacy of the framework and our proposed method outperforms other diffusion-classifiers with significantly fewer time-step samplings.
\end{itemize}

\section{Related Work}
Diffusion models\cite{ho2020denoising, song2020denoising, rombach2022high, karras2022elucidating} have garnered significant interest for their exceptional generative abilities, especially in image generation, where they have outperformed GAN models\cite{goodfellow2020generative} in producing high-quality images. In 2021, OpenAI introduced Classifier Guidance Diffusion\cite{dhariwal2021diffusion}, a technique that facilitates gradient adjustment of images in the generation phase by diffusion models, thus permitting conditional generation based on specified categories. Following this, in 2022, Google unveiled Classifier-Free Guidance Diffusion\cite{ho2022classifier}, a method that eliminates the need for training a separate explicit classifier and instead integrates conditional guidance within the training phase, resulting in enhanced generative capabilities. The latent diffusion model\cite{rombach2022high} maps images to a latent space for both the noising and denoising processes, significantly reducing training expenses and expanding the potential applications and generative capabilities of diffusion models. This approach has laid a crucial foundation for the development of Stable Diffusion, showcasing the versatility and adaptability of diffusion models across various domains \cite{xu2023open, tang2024emergent, zhao2023unleashing}.

In this context, utilizing diffusion models for image classification introduces a novel approach.
SBGC\cite{zimmermann2021score} employs score-based generative models to estimate the conditional likelihood $P(x|y)$, which is then utilized for image classification\cite{li2023your, clark2024text,chen2023robust} and further enhanced by integrating diffusion models into these task.
RDC \cite{chen2023robust} leverages diffusion models to evaluate its adversarial robustness in comparison to traditional discriminative classifiers. Li et al. \cite{li2023your} approximates the log probability $\log p_{\Theta}(x|c)$ through EBLO and identifies the optimal condition $c$ by analyzing noise predictions in the denoising phase, demonstrating notable zero-shot classification capabilities.
Furthermore, a diffusion-based classifier applied to the DiT-XL/2 model \cite{peebles2023scalable}, trained on the ImageNet dataset for supervised classification, achieved a notable accuracy of 79.1\% on the ImageNet validation set. Nevertheless, this performance level does not yet meet the benchmarks established by leading contemporary supervised deep networks on ImageNet, such as ViT-Base, DeiT-Small, and DeiT-Base, all of which can experience significant improvements in classification performance through our methodology. In \cite{clark2024text}, the authors similarly utilize a diffusion model as a classifier. The distinction between these two papers lies that \cite{clark2024text} introduces learnable time-step sampling weights and shifts the loss function focus from measuring the discrepancy between predicted noise and added noise, to quantifying the difference between denoised image and original image.
Despite accelerating the sampling process\cite{li2023your, clark2024text}, the inference duration for a single image from ImageNet can extend to two hours on a GeForce RTX 4090 GPU, which significantly limits the practical application of diffusion models in real-world scenarios despite their innovative use.

In our work, we challenge the conventional belief that ``discriminative models typically perform better in image classification tasks'' by synergizing diffusion models with discriminative models through confidence protector and posterior probability adjustment. This research significantly enhances the accuracy of discriminative models across various architectures, while reducing the inference time to merely 1\% of that of existing diffusion-based classifiers.

\section{Framework}

In this section, we introduce the Diffusion-Based Discriminative Model Enhancement Framework (DBMEF). We first provide an overview of the diffusion model in Sec. \ref{sec:diffusion_model}, and  we introduce the overall process of DBMEF in Sec. \ref{sec:progress}. In Sec. \ref{sec:confidence}, we introduce the confidence protector and diffusion model classifier. In Sec. \ref{sec:negative}, we present two effective methods to further improve the performance. The Pytorch-like pseudo algorithm of DBMEF is given in Appendix. A.

\subsection{Preliminary Knowledge on Diffusion Models}\label{sec:diffusion_model}
The diffusion model \cite{ho2020denoising, song2020denoising, rombach2022high} consists of two processes: forward diffusion with adding noise and inverse diffusion with denoising. In the forward process, a small Gaussian noise is gradually added to a true data distribution $\boldsymbol{x}_0 \sim q(\boldsymbol{x})$. The distribution eventually converges to an isotropic Gaussian distribution. The mean and variance of the added noise are determined by parameter $\beta_t$, then we get:
\begin{equation}
\begin{aligned}
q\left(\boldsymbol{x}_t \mid \boldsymbol{x}_{t-1}\right)&=\mathcal{N}\left(\boldsymbol{x}_t ; \sqrt{1-\beta_t} \boldsymbol{x}_{t-1}, \beta_t I\right)  \\ 
\text { s.t. } & 0<\beta_t<1. 
\end{aligned}
\end{equation}

In the reverse process, the original data is recovered from Gaussian noise $\boldsymbol{x}_T \sim \mathcal{N}(0, \mathbf{I})$. 
In \cite{ho2020denoising}, the posterior probability $P_\theta$ is approximated using a time-conditioned deep model, yielding:
\begin{equation}
\begin{aligned}
P_\theta\left(\boldsymbol{x}_{t-1} \mid \boldsymbol{x}_t\right)&=\mathcal{N}\left(\mu_\theta\left(\boldsymbol{x}_t, t\right), \Sigma_\theta\left(\boldsymbol{x}_t, t\right)\right), \\
P_{\theta}(\boldsymbol{x}_{0:T}) &= p(\boldsymbol{x}_T) \prod_{t=1}^{T} P_{\theta}(\boldsymbol{x}_{t-1} | \boldsymbol{x}_t).
\end{aligned}
\end{equation}

The conditional diffusion model \cite{ho2022classifier} utilize the ELBO as an approximation to the class-conditional log-likelihood as:
\begin{equation}
    \log P_\theta(\boldsymbol{x} \mid y) \geq \mathbb{E}_{q(\boldsymbol{x}_{0:T})} \left[ \log \frac{P_\theta(\boldsymbol{x}_{0:T},y)}{q(\boldsymbol{x}_{1:T} | \boldsymbol{x}_0)} \right].
\label{eq:log}
\end{equation}

By reparameterizing the right half of Eq.~(\ref{eq:log}), similar to \cite{ho2020denoising}, we can obtain the following equivalent relationship:
\begin{equation}
\label{eq:repara}
\begin{aligned}
    \mathbb{E}_{q(\boldsymbol{x}_{0:T})} \left[ \log \frac{P_\theta(\boldsymbol{x}_{0:T},y)}{q(\boldsymbol{x}_{1:T} | \boldsymbol{x}_0)} \right] \iff \\ 
     -\mathbb{E}_{t \sim[1, T], \boldsymbol{x}, \varepsilon_t}\left[\left\|\varepsilon_t-\varepsilon_\theta\left(\boldsymbol{x}_t, t, y\right)\right\|^2\right].
\end{aligned}
\end{equation}

\begin{figure*}[t]
  \centering
  \includegraphics[width=0.8\textwidth]{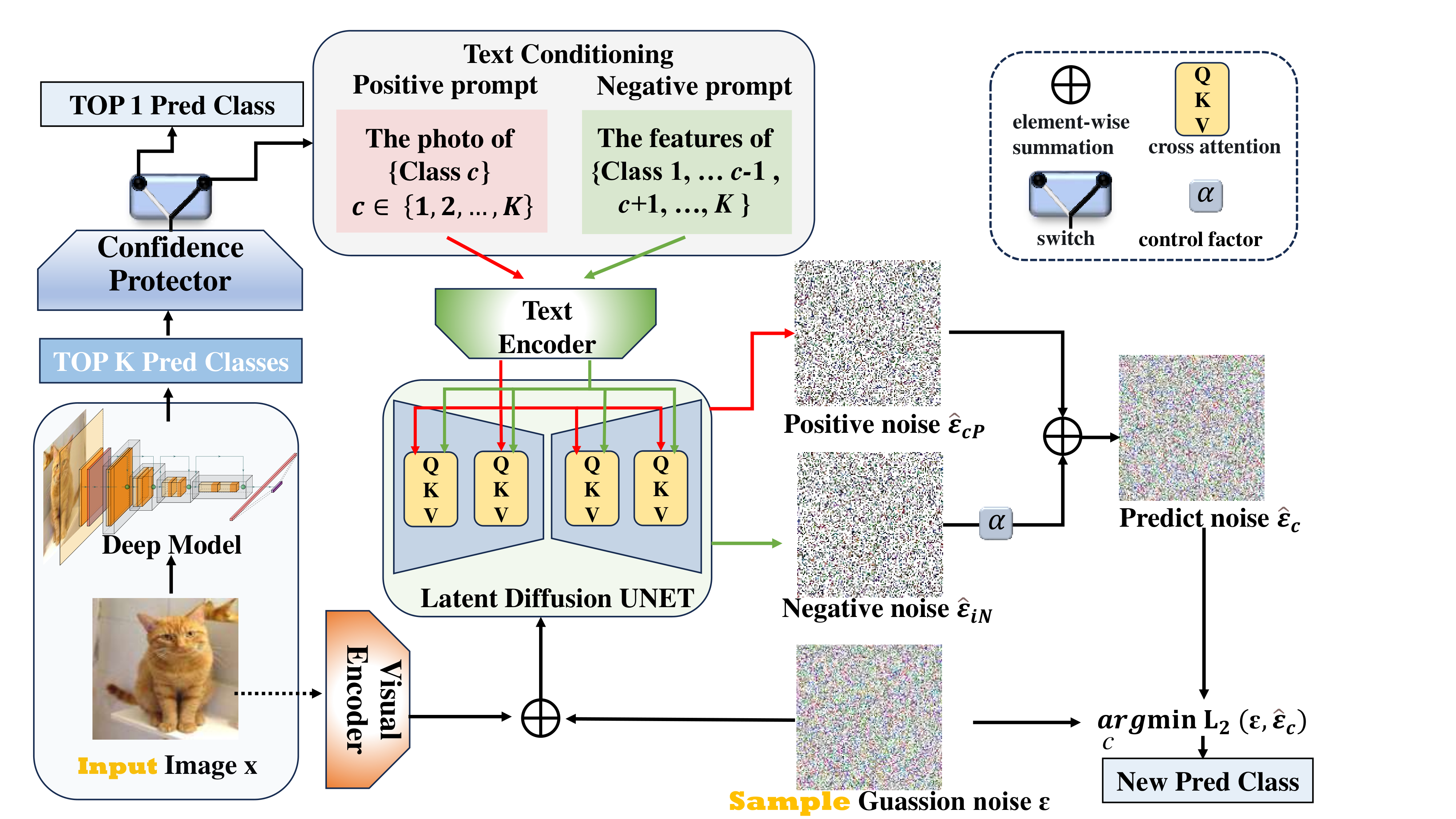}
  \caption{An overview of the Diffusion-Based Discriminative Model Enhancement Framework. For an input image $\boldsymbol{x}$, it first passes through a deep neural network to obtain its top-\(k\) labels. Then, a confidence protector determines the need for further analysis through a diffusion model. If required, positive and negative text conditions, derived from the top-\(k\) labels, are generated. These conditions, alongside 
$\boldsymbol{x}$, are then fed into the diffusion model.  The label with the best denoising outcome is selected as the new predicted label.}
  \label{fig:image1}
\end{figure*}

\subsection{DBMEF Overview} 
\label{sec:progress}
Fig. \ref{fig:image1} illustrates the overall architecture of DBMEF, which comprises two critical components: confidence protector and diffusion model classifier. We further introduce two strategies to enhance the framework's performance: combining positive and negative text conditions and a voting mechanism. The DBMEF process is outlined as follows:
\begin{itemize}
 \item[$\bullet$] An input image 
$\boldsymbol{x}$ is first processed through a deep discriminative model to extract its top-\(k\)  labels. These labels are subsequently inputted into a confidence protector that employs a protection threshold to ascertain the necessity of further re-evaluation via a diffusion model. 

\item[$\bullet$] If the re-evaluation is required, both positive and negative text conditions are generated based on the top-\(k\)  labels. Thereafter, the image 
$\boldsymbol{x}$, in conjunction with these text conditions, is introduced into a diffusion model classifier.

\item[$\bullet$]
 Within this classifier, denoising is conducted under the governance of text condition controls. The denoising outcome is modulated by a negative control factor $\lambda$,
which amalgamates the results from both  positive and negative sets of text conditions to identify the label demonstrating the best predicted noise. 

\item[$\bullet$] 
Based on practical considerations, DBMEF can also flexibly extend from single-sample prediction of noise to multiple-sample prediction, further enhancing its effectiveness through consensus voting on the predictions.
\end{itemize}

\subsection{Confidence Protector and Diffusion Classifier}
\label{sec:confidence}
The utility of the Confidence Protector is to emulate the human brain's determination of confidence when recognizing visual signals. Images that do not evoke enough confidence necessitate rethinking, while those recognized with sufficient assurance bypass the ``guess and verify" process.

For the protection threshold ($Prot$) within the Confidence Protector, we conduct a hypothesis test on the following issue: for an input image $\boldsymbol{x}$, is the output after processing by \(f(\cdot)\) reliable?
\begin{equation}
H_0: f(\boldsymbol{x}) \text{ is reliable} \quad \text{vs} \quad H_1: f(\boldsymbol{x}) \text{ is not reliable}
\end{equation}
Here the null hypothesis $H_0$ implies that the test input $\boldsymbol{x}$ can be reliably classified.

Regarding the determination of whether to reject the null hypothesis ($H_0$), we propose to derive a test statistic based on correctly classified samples within the training set where the discriminative model provides reliable predictions, as suggested in \cite{zhu2022rethinking}. Given an $f(\cdot)$, an input image $\boldsymbol{x}_{k}$, assuming there are $C$ categories, we can formulate the maximum probability of the model output as:
\begin{equation}
\begin{aligned}
S(\boldsymbol{x}_{k})&=\max _{j \in[C]}{\frac{\exp \left(f^j(\boldsymbol{x}_{k}) \right)}{\sum_{i=1}^C \exp \left(f^i(\boldsymbol{x}_{k}) \right)}}.
\end{aligned}
\label{eq:Svalue}
\end{equation}

Then, we can obtain a set of the maximum probability for the correctly classified images $\boldsymbol{S}$: $\left\{S(\boldsymbol{x}_{1}), S(\boldsymbol{x}_{2}),  \ldots, S(\boldsymbol{x}_{N})\right\}$, where $S(\boldsymbol{x}_{i})$ represents the maximum probability of the model output for the $i$-th correctly classified image. We regard the lower \(\alpha\) percentile of $\boldsymbol{S}$ as $S_{\alpha}$ as our test statistic and construct the rejection region for the null hypothesis \(H_0\) as: \begin{equation}
{\mathcal R} = \{\mathbf{x}: S({\boldsymbol{x}}) \leq S_\alpha\},
\end{equation}

For any test input $\boldsymbol{x}$, the null hypothesis is rejected if $\boldsymbol{x} \in \mathcal R.$
It is noteworthy that the \(\alpha\) represents the significance level and also represents the maximum probability of the test making a Type I error which means the mistaken rejection of an actually true null hypothesis $H_0$. Protecting the null hypothesis \(H_0\) is crucial in this hypothesis test, this leads to the introduction of $Prot$:
\begin{equation}
Prot = 1 - \alpha,
\end{equation}
as a crucial parameter for the Confidence Protector.


Therefore, in order to reduce the probability of Type I errors, for preliminary discriminative models already exhibiting high accuracy, a higher $Prot$ is required while for a weaker discriminative model the $Prot$ should be appropriately smaller. In general, the range of $Prot$ is the open interval $(0, 1)$. When $Prot$ is 0, it means that all images need to undergo DBMEF. Conversely, when $Prot = 1$, it signifies that no images require re-thinking via DBMEF.

The diffusion model classifier transforms the conditional denoising outcomes of the diffusion model into an estimation of the posterior probability. In the Diffusion Model Classifier, the  test dataset is set to $\left\{\left(\boldsymbol{x}^1, y^1\right), \ldots, \left(\boldsymbol{x}^n, y^n\right)\right\}$, for a given 
$\boldsymbol{x}$, $y \in \{C_{1},C_{2}, \ldots, C_{K}\}$ represents the top-$K$ possible labels output by the preliminary discriminative model. We transform the posterior probability $P(y=C_i\mid \boldsymbol{x})$ using Bayes' theorem and the reclassification results of the diffusion model are defined as follows:
\begin{equation}
\begin{aligned}
\hat{y} & =\underset{C_i}{\operatorname{argmax}} P_\theta\left(y=C_i \mid \boldsymbol{x}\right) \\
& =\underset{C_i}{\operatorname{argmax}} \frac{P_\theta\left(\boldsymbol{x} \mid y=C_i\right)}{\sum_j P\left(y=C_j\right) P_\theta\left(\boldsymbol{x} \mid y=C_j\right)} P\left(y=C_i\right),
\end{aligned}
\end{equation}
To simplify the problem, we assume equiprobable classes and have $P\left(y=C_i\right)=\frac{1}{K}$. Thus, 
\begin{equation}
    \hat{y}=\underset{C_i}{\operatorname{argmax}} \log P_\theta(\boldsymbol{x} \mid y=C_i).
\label{eq:y}
\end{equation}

Then we use the reparameterized ELBO in Eq. (\ref{eq:repara}) to replace $P_\theta(\boldsymbol{x} \mid y=C_i)$ in Eq. (\ref{eq:y}) as:
\begin{equation}
\label{eq: result}
\begin{aligned}
    \hat{y} & \approx \underset{C_i}{\operatorname{argmax}}
    \mathbb{E}_{q(\boldsymbol{x}_{0:T})} \left[ \log \frac{P_\theta(\boldsymbol{x}_{0:T},y=C_i)}{q(\boldsymbol{x}_{1:T} | \boldsymbol{x}_0)} \right]\\
    &=\underset{C_i}{\operatorname{argmin}} \mathbb{E}_{t \sim[1, T], \boldsymbol{x}, \varepsilon_t}\left[\left\|\varepsilon_t-\varepsilon_\theta\left(\boldsymbol{x}_t, t, C_i\right)\right\|^2\right].
\end{aligned}
\end{equation}

the loss function is usually used to reparameterize the variational lower bound (VLB) as follows:
\begin{equation}
  L_{VLB} = E_{q\left(\boldsymbol{x}_0: T\right)}\left[\log \frac{q\left(\boldsymbol{x}_{1: T} \mid \boldsymbol{x}_0\right)}{P_\theta\left(\boldsymbol{x}_0: T\right)}\right],  
\end{equation}
\begin{equation}
    L_{simple}=E_{t \sim[1, T], \boldsymbol{x}, \varepsilon_t}\left[\left\|\varepsilon_t-\varepsilon_\theta\left(\boldsymbol{x}_t, t, C_i\right)\right\|^2\right].
\end{equation}
Therefore, $\hat{y}$ can be written as:
\begin{equation}
\label{eq: result}
    \begin{aligned}
\hat{y} & \approx \underset{C_i}{\operatorname{argmin}} L_{simple}\left(\boldsymbol{x}, y_{c_i}\right) \\
& =\underset{C_i}{\operatorname{argmin}} E_{t \sim[1, T], \boldsymbol{x}, \varepsilon_t}\left[\left\|\varepsilon_t-\varepsilon_\theta\left(\boldsymbol{x}_t, t, C_i\right)\right\|^2\right].
\end{aligned}
\end{equation}

\subsection{Negative Combination and Voting}
\label{sec:negative}
The negative text conditions represent a reverse selection from the perspective of elimination. For each of the \(k\) alternative classes \( \{C_1, \ldots, C_k\} \) for image $\boldsymbol{x}$, we construct the negative text conditions \(N_i\) for the \(i\)-th alternative class \(C_i\) as follows: \(N_i=\) \(\{\text{The features of } C_1, \cdots, C_{i-1}, C_{i+1}, \cdots, C_k\}\). Under these negative text conditions, the predictive results of the diffusion model classifier in Eq.~(\ref{eq: result}) can be reformulated as:
\begin{equation}
    \hat{y} = \underset{N_i}{\operatorname{argmax}} \mathbb{E}_{t \sim[1, T], \boldsymbol{x}, \varepsilon_t} \left[ \left\| \varepsilon_t - \varepsilon_\theta \left( \boldsymbol{x}_t, t, N_i \right) \right\|^2 \right].
\end{equation}

Negative and positive text conditions serve as dual methods for articulating the same label. Therefore, combining both in a specific ratio is a natural approach. By integrating both, we anticipate an enhancement in label differentiation, which in turn is expected to elevate the model's precision. Following the \cite{dhariwal2021diffusion}, after processing the positive and negative texts through the identical text-encoder, which generates distinct noises, the predicted noise for these divergent text conditions is merged in an element-wise summation at a predetermined ratio named the negative control factor $\lambda$ ,as shown in the equation below:
\begin{equation}
    Noise = Noise_{neg} + \lambda (Noise_{pos} - Noise_{neg}) \quad (\lambda \geq 1).
\end{equation}

Moreover, voting ensemble \cite{dietterich2000ensemble} is a common method to improve model accuracy. In our work, we have chosen a voting method that balances computational resources and performance, selecting the number of models participating in the vote as 5. Each model involved in the voting has the same hyperparameters and is a result of combining positive and negative text conditions within the DBMEF.

\section{Experiments}
\label{exp}
This section provides extensive experiments to answer the following questions:
\begin{itemize}
\item[$\bullet$] Does our framework support various architectures of deep discriminative models with different training methods? See Sec. \ref{main_exp}.

\item[$\bullet$] Does our framework maintain its effectiveness under scenarios of distribution shifts and low-resolution conditions? See Sec. \ref{42} and Sec. \ref{low}.

\item[$\bullet$] What roles do the individual components of the framework play in enhancing model performance and how hyperparameters affect framework performance? See Appendix. B and Appendix. C.
\end{itemize}
\subsection{Performance across Different Models}
\label{main_exp}
\textbf{Baseline:} We choose 17 discriminative models based on various training methods, datasets, and  architectures. The selection comprises nine models employing supervised learning techniques, including DeiT-Base (DeiT-b), DeiT-Small (Deit-s), ViT-Base (ViT-b), ViT-Small (ViT-s), ResNet50, ResNet18, VGG16, MobileNetV3\cite{howard2019searching}, and TinyNet\cite{han2020model}; six models that utilize self-supervised training methods and are fine-tuned on the linear layers with ImageNet2012 training set—specifically ViTb-MAE\cite{he2022masked}, ViTl-MAE, ViTs-DINOv2\cite{oquab2023dinov2}, ViTb-DINOv2 and ResNet50-SimCLR\cite{chen2020simple}, ResNet101-SimCLR; and two models using contrastive learning with fine-tuned linear layers:  ViTb-CLIP\cite{radford2021learning}, and ViTh-CLIP. The pretrained weights of nine supervised backbone models and ViTb-CLIP, ViTh-CLIP, as well as the image preprocessing methods used during inference, were sourced from the TIMM library \footnote{https://github.com/huggingface/pytorch-image-models}. The weights for the MAE model and the DINOv2 model are sourced from the official open-source repositories of Facebook and Google, respectively.

\textbf{Set up details:} Regarding the diffusion model, we adopt the most popular and mature architecture --- Stable Diffusion \text{V1-5}. For the impact of other versions and types of diffusion models on DBMEF, please refer to Appendix. B.
For DeiT-b, ViTh-CLIP, ViTb-CLIP, ViTb-MAE, ViTl-MAE, ViTb-DINOv2, we set $Prot$ at 0.99, while for other models, $Prot$ is set at 0.95. The time steps is set to 30, $\lambda$ is fixed at 1.1, and the number of sub-models participating in voting is set to 5. In the hyperparameter experimentation section(see Appendix. B), we will investigate the impact of these key parameters on the potential for framework effectiveness enhancement.
The evaluation experiments are conducted on the ImageNet2012-1k validation set. We fix the random seed and repeat the process five times, using the average as the reported result.

\begin{table*}[ht]
  \centering
  \setlength{\tabcolsep}{6pt}
  \begin{tabular}[0.8\textwidth]{@{}c|cccccccc@{}}
    \toprule
    Training-method & Model & top1 & w/o p & p$+$neg & p$+$pos & p$+$c & p$+$c$+$v & $\Delta$ \tabularnewline
    \midrule
    \multirow{9}{*}{\makecell{ \bfseries supervised learning \\ \bfseries on backbone model}}
    & \bfseries DeiT-b & 81.98\% & 66.37\% & 82.02\% & 82.41\% & 82.45\% & \bfseries 82.67\% & \bfseries 0.69\% \\
    & \bfseries ViT-b & 80.93\% & 64.98\% & 81.35\% & 81.40\% & 81.43\% & \bfseries 81.73\% & \bfseries 0.80\% \\
    & \bfseries DeiT-s& 79.86\% & 64.88\% & 80.18\% & 80.35\% & 80.42\% & \bfseries 80.78\% & \bfseries 0.92\% \\
    & \bfseries ResNet50 & 76.15\% & 64.39\% & 76.68\% & 77.22\% & 77.40\% & \bfseries 77.66\% & \bfseries 1.51\% \\
    & \bfseries ViT-s & 75.99\% & 63.95\% & 76.72\% & 77.02\% & 77.10\% & \bfseries 77.64\% & \bfseries 1.65\% \\
    & \bfseries VGG16 & 71.58\% & 63.98\% & 72.78\% & 73.39\% & 73.52\% & \bfseries 73.96\% & \bfseries 2.38\% \\
    & \bfseries ResNet18 &  69.76\% &  63.44\% &  71.16\% &  71.85\% &  71.98\% & \bfseries 72.41\% & \bfseries 2.65\% \\
    & \bfseries Mobilenetv3 & 67.64\% & 63.68\% & 69.55\% & 70.27\% &  70.40\% & \bfseries 70.79\% & \bfseries 3.15\% \\
    & \bfseries Tinynet & 66.96\% & 63.44\% & 68.87\% & 69.65\% &  69.72\% & \bfseries 70.23\% & \bfseries 3.27\% \\
    \midrule
    \multirow{6}{*}{\makecell{\bfseries self-supervised learning\\ \bfseries + finetuning}} & \bfseries ViTb-MAE &83.63\% &65.97\% &83.90\% &83.95\% &83.98\% & \bfseries 84.12\% & \bfseries 0.49\% \\
    & \bfseries ViTl-MAE &85.92\% &66.15\% &86.09\% &86.14\% &86.21\% & \bfseries 86.29\% & \bfseries 0.37\% \\
    & \bfseries ViTs-DINOv2 &79.11\% &65.14\% &79.71\% &79.95\% &80.14\% & \bfseries 80.35\% & \bfseries 1.24\% \\
    & \bfseries ViTb-DINOv2 &82.01\% &64.96\% &82.21\% &82.48\% &82.52\% & \bfseries 82.69\% & \bfseries 0.68\% \\
    & \bfseries ResNet50-SimCLR &76.31\% &64.51\% &77.35\% &77.44\% &77.53\% & \bfseries 77.71\% & \bfseries 1.40\%\\
    & \bfseries ResNet101-SimCLR &78.22\% &65.03\% &78.41\% &78.75\% &78.82\% & \bfseries 79.03\% & \bfseries 0.81\% \\
    \midrule
    \multirow{2}{*}{\makecell{\bfseries contrastive learning \\ \bfseries + finetuning}} 
    & \bfseries ViTb-CLIP &85.21\% &65.66\% &85.33\% &85.36\% &85.38\% & \bfseries 85.46\% & \bfseries 0.25\% \\
    & \bfseries ViTh-CLIP & 88.59\% & 65.81\% & 88.60\% & 88.61\% & 88.65\% & \bfseries 88.78\% & \bfseries 0.19\% \\
    \bottomrule
  \end{tabular}
  \caption{The method proposed in this article brings performance gains across different model architectures. We have compared the impact of various strategies for augmenting classification models with generative models. Here, top1 denotes the model's baseline accuracy. w/o p indicates using a strategy similar to that in \cite{li2023your}, where the original task of selecting 1 out of 1000 is simplified to selecting 1 out of 5 without confidence protection. p$+$neg means the model has been processed by the confidence protector and only uses negative text as the text condition. p$+$pos indicates the model has been processed by the confidence protector and only uses positive text as the text condition. p$+$c represents the model being processed by the confidence protector and combining positive and negative text conditions. p$+$c$+$v signifies a voting process involving five p$+$c combinations, that is the most comprehensive DBMEF. $\Delta$ represents the improvement in the original top1 accuracy of the model achieved by applying DBMEF.
}
 \label{tab:model_performance}
\end{table*}
\textbf{Results:} Tab.~\ref{tab:model_performance} presents the performance enhancements of 17 deep discriminative models following their integration with DBMEF. The accuracy improvements on the ImageNet1k validation set ranged from 0.19\% to 3.27\%. Notably, to maintain a balance between accuracy and inference time, all models were configured to 30 timesteps, without optimizing other hyperparameters for optimal settings. Consequently, the full potential of the framework to boost the performance of deep discriminative models may not be entirely captured in these results. Nevertheless, the framework has demonstrated substantial improvements across the board, including for the highly advanced ViTh-CLIP model, which was derived from CLIP and trained and fine-tuned on LAION-2B and ImageNet-12K, further fine-tuned on ImageNet1k training set and the more modest TinyNet, underscoring its universal applicability.

The fourth column(w/o p) of Tab.~\ref{tab:model_performance} displays the results of directly reclassifying the top 5 labels from the discriminative model through the diffusion model without the protection step. This approach is similar to the traditional method of classification using diffusion models, simplifying the selection from 1 out of 1000 to 1 out of 5. Although its accuracy is higher than the performance in the previous works \cite{li2023your,clark2024text} on ImageNet, it is still significantly lower than the accuracy of the discriminative models themselves. This demonstrates that the unprotected method leads to inferior outcomes because re-evaluating all test input images might misclassify some of the results that the preliminary discriminative model had accurately predicted. This action lowers the overall performance of image recognition, thereby illustrating the essential role of confidence protector operations. More intuitive visualization results are presented in Appendix C.

The fifth column(p $+$ neg) of Tab.~\ref{tab:model_performance} shows the results when applying both the Confidence Protector and solely the negative text condition, achieving classification accuracy that surpass the baseline top1 performance. However, its improvement is not as significant as when the Confidence Protector and only positive text condition are applied. The accuracy in sixth column(p+pos) far exceeds that of without protector (w/o p) in the third column, highlighting the crucial role of the Confidence Protector.

In the seventh column(p+c) of Tab.~\ref{tab:model_performance}, the effect achieved by combining both the positive and negative text conditions using $\lambda$ surpasses that of using either individual text condition alone, thereby demonstrating the effectiveness of employing 
$\lambda$ to integrate these conditions. Moreover, as shown in the eighth column(p + c + v) of Tab.~\ref{tab:model_performance}, the classification performance can be further enhanced by applying a voting mechanism to p+c. Ultimately, without additional hyperparameter tuning and only choosing small timesteps, DBMEF enhances the performance of these deep models by 0.19\%-3.01\%, fully demonstrating the excellent performance and potential of our framework.

\subsection{Performance against Distribution Shifts}
\label{42}
\textbf{Baseline:} We use the accuracy of the aforementioned four pre-trained deep discriminative models—ViT-Base, DeiT-Small, ResNet50, VGG16 on ImageNet-S, ImageNet-A, ImageNet-V2, ImageNet-E (background), and ImageNet-E (position) as the baseline.

\textbf{Set up details:} We utilize the complete Diffusion-Based Discriminative Model Enhancement Framework, which includes 5 submodels for voting formed by positive and negative text condition combinations. The diffusion model selected in the framework is Stable Diffusion V1-5, with a $Prot$ 0.95 and time steps 30. $\lambda$ is set to 1.1. The deep discriminative models chosen within the framework include ViT-b, Resnet50, VGG16, and DeiT-s. The pretrained weights are sourced from the TIMM library. We select the ImageNet-A \cite{hendrycks2021natural}, ImageNet-V2\cite{recht2019imagenet}, ImageNet-S\cite{gao2022large}, and Imagenet-E\cite{li2023imagenet} datasets to assess the robustness of our proposed framework against distribution shifts. Additionally, the category labels of these datasets are subsets of the ImageNet1k labels, so no additional training is required.

\begin{table}[ht]
\centering
\setlength{\tabcolsep}{1.75pt}
\fontsize{9pt}{12pt}\selectfont
\begin{tabular}{lcccccc}
\toprule 
Model & S & A & V2 & E-bg & E-pos \\
\midrule
ViT-b & 29.79\% & 27.23\% & 77.23\% & 67.55\% & 75.39\% \\
ViT-b* & \textbf{30.47\%} &\textbf{ 27.41\%} & \textbf{77.37\% }& \textbf{68.61\%} & \textbf{75.92\%} \\ \hline
Resnet50 & 24.08\% & 0.00\% & 72.30\% & 62.22\% & 65.17\% \\
Resnet50* &\textbf{ 25.82\% }& \textbf{3.02\%} & \textbf{72.98\%} & \textbf{63.35\%} & \textbf{65.31\%} \\ \hline
VGG16 & 17.54\% & 2.57\% & 68.14\% & 56.71\% & 61.21\% \\
VGG16* & \textbf{19.70\%} & \textbf{3.84\%} & \textbf{69.29\%} & \textbf{57.92\%} & \textbf{61.92\%} \\ \hline
DeiT-s & 29.42\% & 18.68\% & 76.42\% & 61.14\% & 70.29\% \\
DeiT-s* & \textbf{31.01\%} & \textbf{19.34\%} & \textbf{77.02\%} & \textbf{62.22\%} & \textbf{70.74\%} \\ 
\bottomrule
\end{tabular}
\caption{Results of the Distribution Shift Experiment. S, A and V2 denote ImageNet-S, ImageNet-A and ImageNet-V2 respectively. The asterisk (*) indicates enhanced classifier using our proposed DBMEF, while metrics without an asterisk represent vanilla classifier. The best results are highlighted in Bold.}
\label{tab:distribution_shift}
\end{table}

As shown in Tab.~\ref{tab:distribution_shift}, the DBMEF demonstrates stable improvement capability when facing different types of distribution shift datasets, including real images misclassified by ResNet50, new images with the same ImageNet1k labels, sketch images, and edited versions of original ImageNet images with altered backgrounds and object poses, with the most significant improvements observed for ImageNet-S and ImageNet-A. Notably, on ImageNet-A, DBMEF improved the performance of Resnet50 from 0.0\% to 3.02\%, indicating that DBMEF endowed the model with the ability to reconsider, thereby enhancing accuracy in more adversarial conditions.

\subsection{Classification on Low-Resolution Datasets}
\label{low}
\textbf{Set up details:} We selected the deep discriminative models ResNet18, ResNet34 and ResNet50. Additionally, we chose the CIFAR-10 and CIFAR-100 datasets for our experiments. These two datasets contain 10 and 100 categories, respectively, with each category's images having a resolution of 32 $\times$ 32 pixels\cite{krizhevsky2009learning}.

\textbf{Baseline:} We use the classification accuracy of ResNet18, ResNet34 and ResNet50, which have been fine-tuned 20 epochs on the CIFAR-10 and CIFAR-100 training sets with weights pre-trained on ImageNet, on their respective test sets as the baseline. The parameter settings for the DBMEF are identical to those described in Sec. \ref{42}.

\begin{table}[ht]
\centering
\renewcommand{\arraystretch}{1.} 
\setlength{\tabcolsep}{1.8pt} 
\begin{tabular}{c|c c}
\toprule
DataSet & CIFAR10 & CIFAR100 \\
\midrule
ResNet18   & 95.16\% & 80.94\% \\
ResNet18* & \textbf{95.20\%} & \textbf{81.22\%} \\
\midrule
ResNet34 & 95.41\% & 81.45\% \\
ResNet34* & \textbf{95.44\%} & \textbf{81.85\%} \\
\midrule
ResNet50 & 96.29\% & 83.85\% \\
ResNet50* & \textbf{96.30\%} & \textbf{84.46\%} \\
\midrule
VGG16 & 91.02\% & 70.49\% \\
VGG16* & \textbf{91.68\%} & \textbf{71.42\%} \\
\bottomrule
\end{tabular}
\caption{Model performance on CIFAR10 and CIFAR100 datasets. The asterisk (*) indicates enhanced classifier using our proposed DBMEF, while metrics without an asterisk represent vanilla classifier. The best results are highlighted in bold.}
\label{tab:model_performance3}
\end{table}

In Tab.~\ref{tab:model_performance3}, despite the inherently high accuracy of ResNet18, ResNet34 and ResNet50 on two datasets, the application of our framework yielded further improvements. This demonstrates the effectiveness of strategic coordination between the protective mechanism and the diffusion classifier, even in scenarios where the baseline accuracy is already substantial. Moreover, for low-resolution images, transitioning the original images to a reduced-dimensional latent space (32 $\times$ 32) — as opposed to the larger latent space dimensions (64 $\times$ 64) utilized in the experiments on ImageNet — resulted in enhanced performance and increased processing speed concurrently. A more detailed discussion of inference speed is presented in the Appendix. D.

\section{Conclusion}
In modern image classification tasks, deep learning methods conventionally use either discriminative models or generative models  independently. Our paper draws inspiration from the human brain's coordination of rapid and slow pathways in recognition tasks, proposing a novel framework—the Diffusion-Based Discriminative Model Enhancement Framework (DBMEF). This framework can effectively enhance the classification accuracy and generalization capability of discriminative models in a plug-and-play and training-free manner. We discovered that DBMEF exhibits strong universality, achieving stable performance improvements across 17 common deep discriminative models, including different network architectures and training methods. Additionally, DBMEF still achieves good improvement effects when facing data with distribution shifts and low-resolution data.
Our work fills a gap within the present research field and aims to motivate researchers to further investigate the integration of diffusion models into more downstream applications, combining discriminative and generative modeling principles to fully harness their respective strengths.

\section*{Acknowledgements}
This work was supported by the National Natural Science Foundation of China (No. 12201048).

\bibliography{main}

\begin{thebibliography}{41}
\providecommand{\natexlab}[1]{#1}

\bibitem[{Chen et~al.(2023)Chen, Dong, Wang, Yang, Duan, Su, and Zhu}]{chen2023robust}
Chen, H.; Dong, Y.; Wang, Z.; Yang, X.; Duan, C.; Su, H.; and Zhu, J. 2023.
\newblock Robust Classification via a Single Diffusion Model.
\newblock \emph{arXiv preprint arXiv:2305.15241}.

\bibitem[{Chen et~al.(2020)Chen, Kornblith, Norouzi, and Hinton}]{chen2020simple}
Chen, T.; Kornblith, S.; Norouzi, M.; and Hinton, G. 2020.
\newblock A simple framework for contrastive learning of visual representations.
\newblock In \emph{International conference on machine learning}, 1597--1607. PMLR.

\bibitem[{Chen et~al.(2014)Chen, Akin, Nern, Tsui, Pecot, and Zipursky}]{chen2014cell}
Chen, Y.; Akin, O.; Nern, A.; Tsui, C.~K.; Pecot, M.~Y.; and Zipursky, S.~L. 2014.
\newblock Cell-type-specific labeling of synapses in vivo through synaptic tagging with recombination.
\newblock \emph{Neuron}, 81(2): 280--293.

\bibitem[{Cheng and Greiner(2013)}]{cheng2013comparing}
Cheng, J.; and Greiner, R. 2013.
\newblock Comparing Bayesian network classifiers.
\newblock \emph{arXiv preprint arXiv:1301.6684}.

\bibitem[{Clark and Jaini(2024)}]{clark2024text}
Clark, K.; and Jaini, P. 2024.
\newblock Text-to-Image Diffusion Models are Zero Shot Classifiers.
\newblock \emph{Advances in Neural Information Processing Systems}, 36.

\bibitem[{Dhariwal and Nichol(2021)}]{dhariwal2021diffusion}
Dhariwal, P.; and Nichol, A. 2021.
\newblock Diffusion models beat gans on image synthesis.
\newblock \emph{Advances in neural information processing systems}, 34: 8780--8794.

\bibitem[{Dietterich(2000)}]{dietterich2000ensemble}
Dietterich, T.~G. 2000.
\newblock Ensemble methods in machine learning.
\newblock In \emph{International workshop on multiple classifier systems}, 1--15. Springer.

\bibitem[{Dosovitskiy et~al.(2020)Dosovitskiy, Beyer, Kolesnikov, Weissenborn, Zhai, Unterthiner, Dehghani, Minderer, Heigold, Gelly et~al.}]{dosovitskiy2020image}
Dosovitskiy, A.; Beyer, L.; Kolesnikov, A.; Weissenborn, D.; Zhai, X.; Unterthiner, T.; Dehghani, M.; Minderer, M.; Heigold, G.; Gelly, S.; et~al. 2020.
\newblock An image is worth 16x16 words: Transformers for image recognition at scale.
\newblock \emph{arXiv preprint arXiv:2010.11929}.

\bibitem[{Gao et~al.(2022)Gao, Li, Yang, Cheng, Han, and Torr}]{gao2022large}
Gao, S.; Li, Z.-Y.; Yang, M.-H.; Cheng, M.-M.; Han, J.; and Torr, P. 2022.
\newblock Large-scale unsupervised semantic segmentation.
\newblock \emph{IEEE transactions on pattern analysis and machine intelligence}.

\bibitem[{Goodfellow et~al.(2020)Goodfellow, Pouget-Abadie, Mirza, Xu, Warde-Farley, Ozair, Courville, and Bengio}]{goodfellow2020generative}
Goodfellow, I.; Pouget-Abadie, J.; Mirza, M.; Xu, B.; Warde-Farley, D.; Ozair, S.; Courville, A.; and Bengio, Y. 2020.
\newblock Generative adversarial networks.
\newblock \emph{Communications of the ACM}, 63(11): 139--144.

\bibitem[{Han et~al.(2020)Han, Wang, Zhang, Zhang, Xu, and Zhang}]{han2020model}
Han, K.; Wang, Y.; Zhang, Q.; Zhang, W.; Xu, C.; and Zhang, T. 2020.
\newblock Model rubik’s cube: Twisting resolution, depth and width for tinynets.
\newblock \emph{Advances in Neural Information Processing Systems}, 33: 19353--19364.

\bibitem[{Han, Zheng, and Zhou(2022)}]{han2022card}
Han, X.; Zheng, H.; and Zhou, M. 2022.
\newblock Card: Classification and regression diffusion models.
\newblock \emph{Advances in Neural Information Processing Systems}, 35: 18100--18115.

\bibitem[{He et~al.(2022)He, Chen, Xie, Li, Doll{\'a}r, and Girshick}]{he2022masked}
He, K.; Chen, X.; Xie, S.; Li, Y.; Doll{\'a}r, P.; and Girshick, R. 2022.
\newblock Masked autoencoders are scalable vision learners.
\newblock In \emph{Proceedings of the IEEE/CVF conference on computer vision and pattern recognition}, 16000--16009.

\bibitem[{He et~al.(2016)He, Zhang, Ren, and Sun}]{he2016deep}
He, K.; Zhang, X.; Ren, S.; and Sun, J. 2016.
\newblock Deep residual learning for image recognition.
\newblock In \emph{Proceedings of the IEEE conference on computer vision and pattern recognition}, 770--778.

\bibitem[{Hendrycks et~al.(2021)Hendrycks, Zhao, Basart, Steinhardt, and Song}]{hendrycks2021natural}
Hendrycks, D.; Zhao, K.; Basart, S.; Steinhardt, J.; and Song, D. 2021.
\newblock Natural adversarial examples.
\newblock In \emph{Proceedings of the IEEE/CVF Conference on Computer Vision and Pattern Recognition}, 15262--15271.

\bibitem[{Ho, Jain, and Abbeel(2020)}]{ho2020denoising}
Ho, J.; Jain, A.; and Abbeel, P. 2020.
\newblock Denoising diffusion probabilistic models.
\newblock \emph{Advances in neural information processing systems}, 33: 6840--6851.

\bibitem[{Ho and Salimans(2022)}]{ho2022classifier}
Ho, J.; and Salimans, T. 2022.
\newblock Classifier-free diffusion guidance.
\newblock \emph{arXiv preprint arXiv:2207.12598}.

\bibitem[{Howard et~al.(2019)Howard, Sandler, Chu, Chen, Chen, Tan, Wang, Zhu, Pang, Vasudevan et~al.}]{howard2019searching}
Howard, A.; Sandler, M.; Chu, G.; Chen, L.-C.; Chen, B.; Tan, M.; Wang, W.; Zhu, Y.; Pang, R.; Vasudevan, V.; et~al. 2019.
\newblock Searching for mobilenetv3.
\newblock In \emph{Proceedings of the IEEE/CVF international conference on computer vision}, 1314--1324.

\bibitem[{Karras et~al.(2022)Karras, Aittala, Aila, and Laine}]{karras2022elucidating}
Karras, T.; Aittala, M.; Aila, T.; and Laine, S. 2022.
\newblock Elucidating the design space of diffusion-based generative models.
\newblock \emph{Advances in Neural Information Processing Systems}, 35: 26565--26577.

\bibitem[{Krizhevsky, Hinton et~al.(2009)}]{krizhevsky2009learning}
Krizhevsky, A.; Hinton, G.; et~al. 2009.
\newblock Learning multiple layers of features from tiny images.
\newblock \emph{Master's thesis, University of Tront}.

\bibitem[{LeCun et~al.(1998)LeCun, Bottou, Bengio, and Haffner}]{lecun1998gradient}
LeCun, Y.; Bottou, L.; Bengio, Y.; and Haffner, P. 1998.
\newblock Gradient-based learning applied to document recognition.
\newblock \emph{Proceedings of the IEEE}, 86(11): 2278--2324.

\bibitem[{Li et~al.(2023{\natexlab{a}})Li, Prabhudesai, Duggal, Brown, and Pathak}]{li2023your}
Li, A.~C.; Prabhudesai, M.; Duggal, S.; Brown, E.~L.; and Pathak, D. 2023{\natexlab{a}}.
\newblock Your Diffusion Model is Secretly a Zero-Shot Classifier.
\newblock In \emph{ICML 2023 Workshop on Structured Probabilistic Inference {\&} Generative Modeling}.

\bibitem[{Li et~al.(2023{\natexlab{b}})Li, Chen, Zhu, Wang, Zhang, and Xue}]{li2023imagenet}
Li, X.; Chen, Y.; Zhu, Y.; Wang, S.; Zhang, R.; and Xue, H. 2023{\natexlab{b}}.
\newblock ImageNet-E: Benchmarking Neural Network Robustness via Attribute Editing.
\newblock In \emph{Proceedings of the IEEE/CVF Conference on Computer Vision and Pattern Recognition}, 20371--20381.

\bibitem[{Oquab et~al.(2023)Oquab, Darcet, Moutakanni, Vo, Szafraniec, Khalidov, Fernandez, Haziza, Massa, El-Nouby et~al.}]{oquab2023dinov2}
Oquab, M.; Darcet, T.; Moutakanni, T.; Vo, H.; Szafraniec, M.; Khalidov, V.; Fernandez, P.; Haziza, D.; Massa, F.; El-Nouby, A.; et~al. 2023.
\newblock Dinov2: Learning robust visual features without supervision.
\newblock \emph{arXiv preprint arXiv:2304.07193}.

\bibitem[{Peebles and Xie(2023)}]{peebles2023scalable}
Peebles, W.; and Xie, S. 2023.
\newblock Scalable diffusion models with transformers.
\newblock In \emph{Proceedings of the IEEE/CVF International Conference on Computer Vision}, 4195--4205.

\bibitem[{Radford et~al.(2021)Radford, Kim, Hallacy, Ramesh, Goh, Agarwal, Sastry, Askell, Mishkin, Clark et~al.}]{radford2021learning}
Radford, A.; Kim, J.~W.; Hallacy, C.; Ramesh, A.; Goh, G.; Agarwal, S.; Sastry, G.; Askell, A.; Mishkin, P.; Clark, J.; et~al. 2021.
\newblock Learning transferable visual models from natural language supervision.
\newblock In \emph{International conference on machine learning}, 8748--8763. PMLR.

\bibitem[{Recht et~al.(2019)Recht, Roelofs, Schmidt, and Shankar}]{recht2019imagenet}
Recht, B.; Roelofs, R.; Schmidt, L.; and Shankar, V. 2019.
\newblock Do imagenet classifiers generalize to imagenet?
\newblock In \emph{International conference on machine learning}, 5389--5400. PMLR.

\bibitem[{Rish et~al.(2001)}]{rish2001empirical}
Rish, I.; et~al. 2001.
\newblock An empirical study of the naive Bayes classifier.
\newblock In \emph{IJCAI 2001 workshop on empirical methods in artificial intelligence}, volume~3, 41--46.

\bibitem[{Rombach et~al.(2022)Rombach, Blattmann, Lorenz, Esser, and Ommer}]{rombach2022high}
Rombach, R.; Blattmann, A.; Lorenz, D.; Esser, P.; and Ommer, B. 2022.
\newblock High-resolution image synthesis with latent diffusion models.
\newblock In \emph{Proceedings of the IEEE/CVF conference on computer vision and pattern recognition}, 10684--10695.

\bibitem[{Sillito, Cudeiro, and Jones(2006)}]{sillito2006always}
Sillito, A.~M.; Cudeiro, J.; and Jones, H.~E. 2006.
\newblock Always returning: feedback and sensory processing in visual cortex and thalamus.
\newblock \emph{Trends in neurosciences}, 29(6): 307--316.

\bibitem[{Simonyan and Zisserman(2014)}]{simonyan2014very}
Simonyan, K.; and Zisserman, A. 2014.
\newblock Very deep convolutional networks for large-scale image recognition.
\newblock \emph{arXiv preprint arXiv:1409.1556}.

\bibitem[{Song, Meng, and Ermon(2020)}]{song2020denoising}
Song, J.; Meng, C.; and Ermon, S. 2020.
\newblock Denoising diffusion implicit models.
\newblock \emph{arXiv preprint arXiv:2010.02502}.

\bibitem[{Song et~al.(2020)Song, Sohl-Dickstein, Kingma, Kumar, Ermon, and Poole}]{song2020score}
Song, Y.; Sohl-Dickstein, J.; Kingma, D.~P.; Kumar, A.; Ermon, S.; and Poole, B. 2020.
\newblock Score-based generative modeling through stochastic differential equations.
\newblock \emph{arXiv preprint arXiv:2011.13456}.

\bibitem[{Tang et~al.(2024)Tang, Jia, Wang, Phoo, and Hariharan}]{tang2024emergent}
Tang, L.; Jia, M.; Wang, Q.; Phoo, C.~P.; and Hariharan, B. 2024.
\newblock Emergent correspondence from image diffusion.
\newblock \emph{Advances in Neural Information Processing Systems}, 36.

\bibitem[{Touvron et~al.(2021)Touvron, Cord, Douze, Massa, Sablayrolles, and J{\'e}gou}]{touvron2021training}
Touvron, H.; Cord, M.; Douze, M.; Massa, F.; Sablayrolles, A.; and J{\'e}gou, H. 2021.
\newblock Training data-efficient image transformers \& distillation through attention.
\newblock In \emph{International conference on machine learning}, 10347--10357. PMLR.

\bibitem[{Xu et~al.(2023)Xu, Liu, Vahdat, Byeon, Wang, and De~Mello}]{xu2023open}
Xu, J.; Liu, S.; Vahdat, A.; Byeon, W.; Wang, X.; and De~Mello, S. 2023.
\newblock Open-vocabulary panoptic segmentation with text-to-image diffusion models.
\newblock In \emph{Proceedings of the IEEE/CVF Conference on Computer Vision and Pattern Recognition}, 2955--2966.

\bibitem[{Yoon, Hwang, and Lee(2021)}]{yoon2021adversarial}
Yoon, J.; Hwang, S.~J.; and Lee, J. 2021.
\newblock Adversarial purification with score-based generative models.
\newblock In \emph{International Conference on Machine Learning}, 12062--12072. PMLR.

\bibitem[{Zhao, Jacobsen, and Grathwohl(2020)}]{zhao2020joint}
Zhao, S.; Jacobsen, J.-H.; and Grathwohl, W. 2020.
\newblock Joint energy-based models for semi-supervised classification.
\newblock In \emph{ICML 2020 Workshop on Uncertainty and Robustness in Deep Learning}, volume~1.

\bibitem[{Zhao et~al.(2023)Zhao, Rao, Liu, Liu, Zhou, and Lu}]{zhao2023unleashing}
Zhao, W.; Rao, Y.; Liu, Z.; Liu, B.; Zhou, J.; and Lu, J. 2023.
\newblock Unleashing text-to-image diffusion models for visual perception.
\newblock \emph{arXiv preprint arXiv:2303.02153}.

\bibitem[{Zhu et~al.(2022)Zhu, Chen, Li, Zhang, Xue, Tian, Jiang, Zheng, and Chen}]{zhu2022rethinking}
Zhu, Y.; Chen, Y.; Li, X.; Zhang, R.; Xue, H.; Tian, X.; Jiang, R.; Zheng, B.; and Chen, Y. 2022.
\newblock Rethinking Out-of-Distribution Detection From a Human-Centric Perspective.
\newblock \emph{arXiv preprint arXiv:2211.16778}.

\bibitem[{Zimmermann et~al.(2021)Zimmermann, Schott, Song, Dunn, and Klindt}]{zimmermann2021score}
Zimmermann, R.~S.; Schott, L.; Song, Y.; Dunn, B.~A.; and Klindt, D.~A. 2021.
\newblock Score-Based Generative Classifiers.
\newblock In \emph{NeurIPS 2021 Workshop on Deep Generative Models and Downstream Applications}.

\end{thebibliography}

\appendix
\clearpage
\newpage
\section{Algorithm}
The PyTorch-like pseudo algorithm of DBMEF is provided in Alg.~\ref{alg:dbmef}. The negative control factor \(\lambda\) modulates the application of text conditions in DBMEF: \(\lambda = 0\) signifies solely utilizing negative text condition in the proposed framework. Conversely, \(\lambda = 1\) corresponds to the utilizing of positive text condition in DBMEF. For \(\lambda > 1\), both positive and negative text conditions are combined in DBMEF. Based on experimental results discussed in the main text, a \(\lambda\) value of 1.1 is advocated.

\begin{algorithm}
\caption{DBMEF}
\label{alg:dbmef}
\begin{algorithmic}[1] 

\STATE \textbf{Input:} deep model $f(\cdot)$, $prot$, Error\_list $[\ ]$, $\lambda$,  test set $\{\boldsymbol{x}^1, \cdots, \boldsymbol{x}^N\}$, timesteps: $\{t_1, t_2, \cdots, t_T\}$
\FOR{$\boldsymbol{x}$ in test set}
    \IF {$\max(\operatorname{softmax}(f(\boldsymbol{x})))<prot$}
        \STATE $K_{class} =\operatorname{topk}(f(\boldsymbol{x}))=\{C_1, \ldots, C_K\}$
        \FOR{$t_i$ in $\{t_1, \ldots, t_T\}$}
            \STATE $\boldsymbol{x}_{t_i}=\sqrt{\bar{\alpha} t_i} \boldsymbol{x} + \sqrt{1-\bar{\alpha_i}}\varepsilon$, $\varepsilon \sim N(0, I)$
            \FOR{$C_j$ in $K_{class}$}
                \STATE $pos = \{C_j\}$
                \STATE $neg = \{C_1,\ldots,C_{j-1},C_{j+1},\ldots,C_K\}$
                \STATE\begin{equation*}
                    \begin{aligned}
                pred_{combine} &= {\varepsilon_{\theta}(\boldsymbol{x}_{t_i}, t_i, neg)}  +\lambda[\varepsilon_{\theta}(\\ &\boldsymbol{x}_{t_i}, t_i, pos)-\varepsilon_{\theta}(\boldsymbol{x}_{t_i}, t_i, neg)]
                \end{aligned}
                \end{equation*} 
                \STATE $Error\_list[C_j].append \left(\|\varepsilon -pred_{combine}\|^2\right) $
            \ENDFOR
        \ENDFOR
        \STATE \textbf{return} $\underset{c_j \in K_{class}}{\operatorname{argmin}} \operatorname{mean}(Error\_list[c_j])$
    \ELSE
    \STATE \textbf{return} $top1$ $f(\boldsymbol{x})$
    \ENDIF
\ENDFOR
\end{algorithmic}
\label{alg}
\end{algorithm}

\section{Hyperparameter Experiments}
In this section, we will verify the impact of different hyperparameters, including $Prot$, timesteps $T$, negative control factor $\lambda$ and the choice of the diffusion model.

\subsection{The impact of Hyperparameters}
\begin{figure*}[t]
  \centering
  \includegraphics[width=1.2\columnwidth]{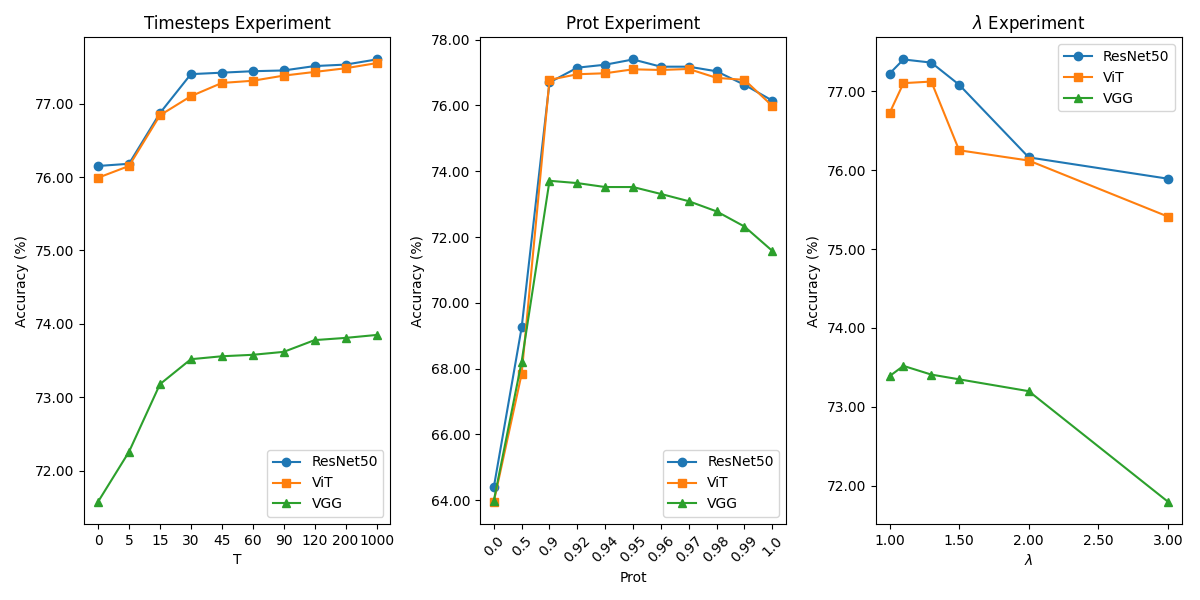}
  \caption{The impact of varying hyperparameters on the accuracy across different deep neural networks. From left to right, the graphs present the relationship between each hyperparameter and the classification accuracy achieved by the respective models, where the point corresponding to $T=0$, $Prot =1.0$ represents the top1 accuracy of the original model and $\lambda = 1.00$ represents the Confidence Protector and only positive text condition are applied in DBMEF.}
  \label{Hyperparameter}
\end{figure*}

For timesteps $T$, we experimented with 9 different time steps ranging from 5 to 1000.
We conducted experiments under 11 different levels of protection threshold and experimented with negative factor $\lambda$ values of 1.0, 1.1, 1.3, 1.5, 2.0 and 3.0. The results are shown in Fig.~\ref{Hyperparameter} and Tab. \ref{tab:performanceT}, \ref{tab:performance2}, \ref{tab:performance3}. 

The results regarding time steps are intuitive; the higher the time step, the higher the accuracy, with the incremental improvement diminishing as the time step increases. Surprisingly, even with a minimal time step of 5, there is still a positive improvement for the three discriminative  models in Tab. \ref{tab:performanceT}. Compared to the 1000 time steps selected for classification using diffusion models directly, our DBMEF significantly increases efficiency while ensuring improved performance. Our method requires a minimum of only $K(5) \times T(5)$ operations, compared to the original articles' $K(1000) \times T(1000)$ operations per image.

The experimental results on $Prot$ show that not protecting the samples with high confidence leads to a significant time consumption, and the accuracy is also lower than that of discriminative models. Overprotecting the samples, however, also results in a decrease in accuracy. In Tab \ref{tab:performance2}, different preliminary discriminative models require varying degrees of protection intensity. Typically, \(Prot\) \(\geq 0.9\) can effectively safeguard the correct samples. However, we recommend that for models with weaker classification performance, such as VGG16, a lower \(Prot\) should be chosen, as rethinking helps to correct samples misclassified by the model. For instance, choosing \(Prot = 0.90\) performs \(0.62\%\) better than \(Prot = 0.97\). On the other hand, for models with relatively good classification performance, we suggest opting for a higher \(Prot\) to protect the initially correctly classified samples more. For example, on ViT-s, choosing \(Prot = 0.97\) yields a \(0.33\%\) better performance than \(Prot = 0.90\). Overall, $Prot$ directly dictates how many samples need to undergo re-correction through the diffusion classification process. Both overly low and excessively high levels of $Prot$ can reduce the final classification results. From the angle of human image cognition, not all images necessitate a ``guess-verify" cognitive chain. For simpler images, identification can be swiftly accomplished within the brain's rapid pathway, with the confidence protection threshold quantitatively mimicking the brain's threshold for whether a visual signal requires re-evaluation.

\begin{figure}[t]
  \centering
  \includegraphics[width=0.9\columnwidth]{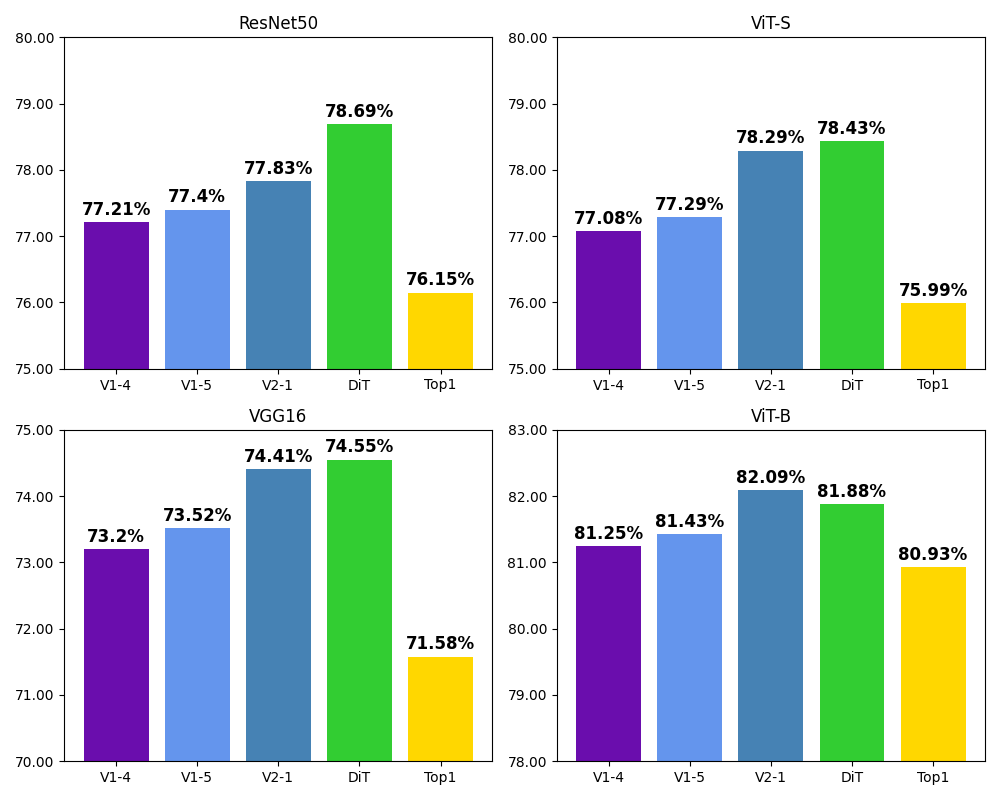}
  \caption{Hyperparameter(diffusion models) results. The illustration of the impact of different diffusion models on classification accuracy across different discriminative models.}
  \label{Hyperparameter1}
\end{figure}

The model performance first increases and then decreases with the negative control factor $\lambda$ in Tab. \ref{tab:performance3}, suggesting that when the intensity of the negative signal is too strong, it does not help the model eliminate incorrect answers but instead causes interference. In scenarios where only negative conditional text is present, it essentially entails reverse selection for images with uncertain labels. When both positive and negative samples are present, the \(\lambda\) factor dictates the influence strength of negative conditional text. Experiments have shown that a too large \(\lambda\) value can decrease the model's classification precision, whereas an appropriate \(\lambda\) can efficiently amalgamate the insights drawn from both positive and negative text conditions to aid the model in making superior choices. This mechanism mirrors the push-pull feedback principle observed in human cognition, where the existence of negative samples serves to accentuate the distinctions between categories prone to confusion, thereby improving the framework's efficacy.

\begin{table*}[htbp]
  \centering
  \renewcommand{\arraystretch}{1.0}
  \begin{tabular}{lcccccccccc}
    \toprule
    Model & 0 & 5 & 15 & 30 & 45 & 60 & 90 & 120 & 200 & 1000 \\
    \midrule
    Rs50 & 76.15\% & 76.18\% & 76.87\% & 77.40\% & 77.42\% & 77.44\% & 77.45\% & 77.51\% & 77.53\% & \textbf{77.60\%} \\
    ViT-s & 75.99\% & 76.15\% & 76.84\% & 77.10\% & 77.28\% & 77.31\% & 77.38\% & 77.43\% & 77.48\% & \textbf{77.55\%}\\
    VGG16 & 71.58\% & 72.26\% & 73.18\% & 73.52\% & 73.56\% & 73.58\% & 73.62\% & 73.78\% & 73.81\% & \textbf{73.85\%} \\
    \bottomrule
  \end{tabular}%
  \caption{The performance of DBMEF under different timesteps, where \textit{timesteps} \(= 0\) represents the original classification accuracy of the preliminary discriminative model. At this point, the applied DBMEF is \textit{p + c}, without voting. The best results are highlighted in Bold. 
}
  \label{tab:performanceT}
\end{table*}

\begin{table*}[htbp]
  \centering
  \setlength{\tabcolsep}{4pt}
  \renewcommand{\arraystretch}{1.2}
  \begin{tabular}[0.8\textwidth]{l*{11}{c}}
    \toprule
    Model & 0.00 & 0.50 & 0.90 & 0.92 & 0.94 & 0.95 & 0.96 & 0.97 & 0.98 & 0.99 & 1.00 \\
    \midrule
    Rs50 & 64.39\% & 69.25\% & 76.70\% & 77.15\% & 77.24\% & \textbf{77.40\%} & 77.18\% & 77.18\% & 77.04\% & 76.63\% & 76.15\% \\
    ViT-s & 63.95\% & 67.84\% & 76.78\% & 76.95\% & 76.98\% & 77.10\% & 77.08\% & \textbf{77.11\%} & 76.84\% & 76.78\% & 75.99\% \\
    VGG16 & 63.98\% & 68.21\% & \textbf{73.71\%} & 73.64\% & 73.52\% & 73.52\% & 73.31\% & 73.09\% & 72.78\% & 72.32\% & 71.58\% \\
    \bottomrule
  \end{tabular}%
  \caption{The performance of DBMEF under different values of \(prot\), where \(Prot = 1\) means no images are reclassified, representing the original classification accuracy of the preliminary discriminative model. Conversely, \(Prot = 0\) implies that all images undergo reclassification, akin to the approach of traditional diffusion classifiers. During this experiment, the applied DBMEF is \(prot + pos\), without integrating negative textual conditions and voting. The best results are highlighted in Bold. 
}
  \label{tab:performance2}
\end{table*}

\begin{table*}[ht]
  \centering
  \begin{tabular}[0.8\textwidth]{l*{6}{c}}
    \toprule
    Model & 1.0 & 1.1 & 1.3 & 1.5 & 2.0 & 3.0 \\
    \midrule
    ResNet50 & 77.22\% & \textbf{77.40\%} & 77.36\% & 77.08\% & 76.16\% & 75.89\% \\
    ViT-s & 76.72\% & 77.10\% & \textbf{77.12\%} & 76.25\% & 76.12\% & 75.41\% \\
    VGG16 & 73.39\% & \textbf{73.52\%} & 73.41\% & 73.35\% & 73.20\% & 71.80\% \\
    \bottomrule
  \end{tabular}%
  \caption{The performance of DBMEF under different values of \(\lambda\), where \(\lambda = 1\) represents the DBMEF using the Confidence Protector and solely positive textual conditions. Meanwhile, \(\lambda = 1.1\) is the parameter used in the fifth column of Tab. 1 in main text, denoted as \(p+c\). The best results are highlighted in Bold.
}
  \label{tab:performance3}
\end{table*}

\subsection{The impact of Diffusion Models}
For the selection of diffusion models, we chose Stable Diffusion V1-4, V1-5, V2-1, and DiT. During the experiments, we fixed the timesteps to 30 and $\lambda$ to 1.1. It is important to note that DiT does not have a text-encoder part, so the condition input is the class index of ImageNet. 

As shown in Fig.~\ref{Hyperparameter1}, when employing various diffusion models in the DBMEF framework, although there exist differences in the final classification accuracy, they all notably surpass the baseline Top1 performance, with the DiT model demonstrating the most outstanding performance among them. The DiT used here is based on weights open-sourced by Facebook, trained only on the ImageNet. Our goal is to propose a more universal enhancement framework rather than solely for specific dataset; therefore, the more universally applicable SD is chosen in our main experiment. Intriguingly, as for the three different versions of Stable Diffusion in Tab. \ref{tab:performance4}, newer versions of diffusion have brought more noticeable improvements. When the preceding discriminative model is ViT-s, the improvement effect between V2-1 and V1-4 differs by 1.21\%. Therefore, we believe that with the introduction of newer and more powerful diffusion models, DBMEF will bring about more significant improvements.

\section{Role of Confidence Protector and Diffusion model}
\subsection{Confidence Protector}
In this section, we provide a detailed visualization and analysis of the impact of $Prot$ on the performance of DBMEF. We can categorize the reclassification results of DBMEF into the following four scenarios: the pre-discriminative model correctly classifies $\Rightarrow$ DBMEF correctly reclassifies $(T\_T)$; the pre-discriminative model correctly classifies $\Rightarrow$ DBMEF incorrectly reclassifies $(T\_F)$; the pre-discriminative model incorrectly classifies $\Rightarrow$ DBMEF correctly reclassifies $(F\_T)$; the pre-discriminative model incorrectly classifies $\Rightarrow$ DBMEF incorrectly reclassifies $(F\_F)$; among these, the quantities of $F\_F$ and $T\_T$ do not impact the framework's performance. The accuracy improvement provided by DBMEF can be calculated as:
\begin{equation}
    \Delta = \frac{F\_T - T\_F}{N} \times 100\%,
\end{equation}
where $N$ represents the number of images in the test set.

It's not difficult to see from Tab. \ref{tab: woprofermance} that compared to \textit{w/o p}, \textit{prot + pos} not only significantly improves classification accuracy but also remarkably saves computational time. This underscores the vital role of the Confidence Protector. The presence or absence of the Confidence Protector has a more substantial impact on models with initially high classification accuracy, with differences in classification accuracy reaching up to \(16.42\%\) and \(15.47\%\) for ViT-b and Deit-s, respectively. For models with lower initial classification accuracy, like Resnet18, the difference in classification accuracy with and without the Confidence Protector can be as high as \(8.41\%\). The Confidence Protector enhances model performance by safeguarding samples correctly classified by the preliminary discriminative model, thereby reducing false negatives.

\begin{table}[htbp]
  \centering
  \renewcommand{\arraystretch}{1.5}
  \setlength{\tabcolsep}{5pt} 
    \begin{tabular}{cccccc}
    \toprule
    Model & Total & $T\_F$ & $F\_T$ & $\Delta$ \\
    \midrule
    ViT-b(w/o p) & 50000 & 10671 & 2696 & -15.95\% \\
    ViT-b(p + pos) & 3462 & 376 & 611 & \textbf{0.47\%} \\\hline
    DeiT-s(w/o p) & 50000 & 9928 & 2437 & -14.98\% \\
    DeiT-s(p + pos) & 2782 & 433 & 680  & \textbf{0.49\%} \\\hline
    RN50(w/o p) & 50000 & 9498 & 3621 &  -11.76\% \\
    RN50(p + pos) & 8072 & 1101 & 1638 & \textbf{1.07\%} \\\hline
    ViT-s(w/o p) & 50000 & 9652 & 3632  & -12.04\% \\
    ViT-s(p + pos) & 8134 & 1215 & 1782 & \textbf{1.11\%} \\\hline
    VGG16(w/o p) & 50000 & 8523 & 4723 & -7.60\% \\
    VGG16(p + pos) & 8026 & 807 & 1713 & \textbf{1.81\%} \\\hline
    RN18(w/o p) & 50000 & 7573 & 4377 & -6.32\% \\
    RN18(p + pos) & 8284 & 718 & 1766 & \textbf{2.09\%} \\
    \bottomrule
    \end{tabular}%
    \caption{The comparison of performance with and without the Confidence Protector, where \textit{w/o p} represents the diffusion model classifier with only positive text conditions, \textit{prot $+$ pos} refers to DBEMF incorporating both the Confidence Protector and positive textual conditions. ``Total" indicates the number of images that require reclassification,``Times" denotes the time needed to process all images in the validation set on a Geforce RTX 4090, and $\Delta$ represents the variation in classification performance after applying DBMEF compared to the initial $Top$ 1 outcomes. ResNet18 and ResNet50 are respectively abbreviated as RN18 and RN50.}
    \label{tab: woprofermance}
\end{table}    

\begin{figure*}[h]
  \centering
  \includegraphics[width=1.2\columnwidth]{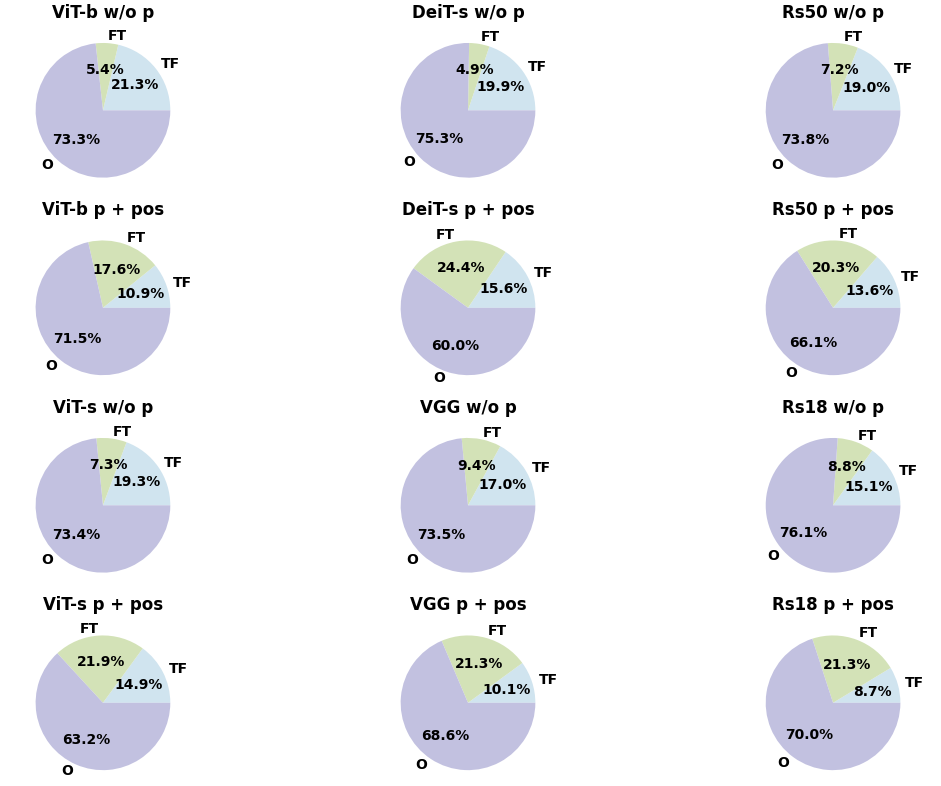}
  \caption{The pie chart illustrating the impact of the presence or absence of the Confidence Protector on image reclassification. Here,``\textbf{O}" represents the total of $T\_T$ and $F\_F$.}
  \label{fig:app1}
\end{figure*}

Furthermore, we visualize the impact of the Confidence Protector on six different preliminary discriminative models with and without it through pie charts in Fig. \ref{fig:app1}, offering a more intuitive understanding of its crucial role. It is evident that the Confidence Protector significantly reduces the proportion of $(T\_F)$ while increasing the proportion of $(F\_T)$. This indicates that the Confidence Protector can effectively simulate the human brain's image recognition process, where only visual signals with high uncertainty undergo a ``guess-verify" re-evaluation pathway.

\begin{table}[ht]
  \centering
  \setlength{\tabcolsep}{3.5pt}
\renewcommand{\arraystretch}{1.0}
  \begin{tabular}{l*{5}{c}}
    \toprule
    Model & V1-4 & V1-5 & V2-1 & DiT & Top1 \\
    \midrule
    ResNet50 & 77.21\% & 77.40\% & 77.83\% & \textbf{78.69\%} & 76.15\% \\
    ViT-s & 77.08\% & 77.29\% & 78.29\% & \textbf{78.43\%} & 75.99\% \\
    VGG16 & 73.20\% & 73.52\% & 74.41\% & \textbf{74.55\%} & 71.58\% \\
    ViT-b & 81.25\% & 81.43\% & \textbf{82.09\%} & 81.88\% & 80.93\% \\
    \bottomrule
  \end{tabular}
  \caption{The performance of DBMEF under different diffusion models acting as the diffusion classifier, with \(timesteps = 30\), \(\lambda = 1.1\), and \(Prot = 0.99\) for ViT-b. For the other three neural networks, \(Prot\) is set to \(0.95\). No voting mechanism was applied in this experiment. The best results are highlighted in Bold.
}
  \label{tab:performance4}
\end{table}

\subsection{The Role of Diffusion Model}
In this section, we will demonstrate through experiments the critical role of the diffusion model in DBMEF. The improvements brought by DBMEF are not due to the reclassification stage using a model that has seen more data, but rather are attributable to the characteristics of the generative model.

In our experimental setup, we substituted the diffusion model within the DBMEF with several highly capable discriminative models, including ResNet101 trained and fine-tuned on ImageNet21k-1k, ViTb-MAE, ViTb-DINOV2, and ViTb-CLIP, each fine-tuned on ImageNet1k. As shown in Tab. \ref{tab:roleofdif}, diffusion models such as SD1-5 and SD2-1 demonstrate superior enhancement effects compared to these alternatives. Upon substituting the second-stage model with other discriminative models familiar with large-scale data, we observed an improvement in performance; however, it was less pronounced than that achieved with our diffusion model configuration. Crucially, the results indicate that deploying more potent discriminative models within the DBMEF does not yield the same level of accuracy as using the second-stage discriminative model in isolation. In contrast, the implementation of diffusion models within DBMEF not only surpasses the performance of standalone discriminative or diffusion models but also underscores the pivotal role of diffusion models in DBMEF.

\begin{table*}[ht]

  \centering
  \setlength{\tabcolsep}{6pt}
  \renewcommand{\arraystretch}{1.0}
    \caption{Performance Comparison, TOP1-1 denotes the first stage accuracy, TOP2-1 denotes the second stage accuracy. TOP1* denotes the final accuracy after enhancement. $\Delta$ represents the difference between the final accuracy and the maximum accuracy achieved in the first and second stages.}
  
  \begin{tabular}{@{}c|ccccc@{}}
    \toprule
    First stage & Second stage & TOP1-1 & TOP1-2 &TOP1* & $\Delta$ \\
    \midrule
    \multirow{6}{*}{ResNet50} \ & ResNet101 & 76.15\% & 79.31\% & 76.98\% & -2.33\% \\
     & ViTb-CLIP & 76.15\% &85.21\%& 77.04\% & -8.17\% \\
     & ViTb-MAE & 76.15\% &83.63\% & 76.42\% & -7.21\% \\
     & ViTb-DINOV2 & 76.15\% &82.01\% & 77.28\% & -4.73\% \\
     & SD1-5 & 76.15\% &64.39\% & 77.40\% & \textbf{1.25\%} \\
     & SD2-1 & 76.15\% &65.72\% & 77.83\% & \textbf{1.68\%} \\
    \bottomrule
  \end{tabular}
  \label{tab:roleofdif}
\end{table*}

\section{Inference Speed Analysis}
We set $Prot$ to $0.95$ and compare the inference speeds with number of timesteps set to $0$, $5$, $10$, and $30$, $0$ represents the original speed of ResNet50, compute the average runtime over three trials on an RTX 4090 GPU on ImageNet valid dataset with pytorch2-1 and batch size 16. In addition, we would like to emphasize that nums of timesteps of 5 can also bring an increase in accuracy to the discriminant model in Tab. \ref{tab:performanceT}. Although DBMEF requires more time compared to solely using discriminative models, it only needs 1\% of the time required by generative models when used alone for classification tasks. Moreover, attempting to improve the accuracy more than 1\% on ImageNet with the same network architecture through methods like expanding the training set and fine-tuning, would necessitate a considerably longer training period. However, our DBMEF offers the advantages of being both plug-and-play and training-free, streamlining the process significantly.
\begin{tabular}{@{}lcccc@{}} 
\toprule
\renewcommand{\arraystretch}{0.5} 
Number of timesteps & 0 & 5 & 10 & 30 \\
\midrule
Times / hours & 0.209 & 0.214 & 0.227 & 0.252 \\
\bottomrule
\end{tabular}

\section{Design Philosophy of Protection Intensity}

In the main text, we have conceptualized the Confidence Protector through the lens of hypothesis testing. This section is dedicated to an in-depth discussion on the formulation of the test statistic within the Confidence Protector framework. The intent behind the Confidence Protector is to discern images that the preliminary discriminative model classifies with assured confidence from those it does not, thereby safeguarding the images that are correctly classified by the model. The essence of hypothesis testing involves utilizing the information from a known sample to infer attributes about the population. When it comes to constructing the test statistic, one typically commences from the standpoint of the null hypothesis (\(H_0\)) to infer, indirectly, the validity of the alternative hypothesis (\(H_1\)). The underlying principle here is that events of low probability are highly unlikely to occur; thus, we calculate the test statistic under the assumption that the null hypothesis holds.

When we return to the specific issue of our test, if the null hypothesis is valid, the output of \(f(\boldsymbol{x})\) is deemed reliable. It follows, then, that selecting the maximum probability of the model's output, \(S(\boldsymbol{x})\), delineated in Eq. 6 in main text, serves as a fitting test statistic. Commencing with the known information (all samples in the training set) prompts us to confront an emerging question: Is it appropriate to compute the statistic \(S\) using the entirety of the training samples?

We posit that computing the statistic based on the samples correctly classified within the training set is more logical. The preliminary discriminative model categorizes all training samples into two cohorts: those correctly classified and those misclassified. Subsequently, we employ the independent samples Mann-Whitney U test to examine the presence of significant disparities between these two groups. The outcomes of these tests for various preliminary discriminative models, such as ResNet50, ResNet18, VGG16 and ViT-b, are detailed in Tab. \ref{tab:statistic}.

\begin{table}[htbp]
\centering
\setlength{\tabcolsep}{1pt}
\renewcommand{\arraystretch}{1.0}
\begin{tabular}[0.9\columnwidth]{@{}ccccc@{}}
\toprule
Model   & Correct &  Misclassified & P-value & Cohen's d \\ \midrule
ResNet50 & 0.980 ± 0.172         & 0.514 ± 0.226      & \textbf{0.000\textsuperscript{***}} & 1.830      \\
ResNet18 & 0.945 ± 0.209        & 0.434 ± 0.232      & \textbf{0.000\textsuperscript{***}} & 1.721     \\
VGG16    & 0.957 ± 0.202        & 0.420 ± 0.224       & \textbf{0.000\textsuperscript{***}} & 1.937
\\
ViT-b    & 0.994 ± 0.121        & 0.594 ± 0.218       & \textbf{0.000\textsuperscript{***}} & 1.995
\\ \bottomrule
\end{tabular}
\caption{The results of the independent samples Mann-Whitney U test, where ``Correct" represents the \(S\) value of samples that were correctly classified in the training set, and ``Misclassified" represents the \(S\) value of samples that were misclassified in the training set. The data shown in the second and third columns of the table are the median ± standard deviation. The P-values are highlighted in Bold.}
\label{tab:statistic}
\end{table}

For the four different preliminary discriminative models, the P-values all demonstrate strong significance. This indicates a significant difference in the \(S\) values between samples correctly classified and those misclassified within the training set. Additionally, the size of Cohen's \(d\) values reveals that the magnitude of difference between the two groups is very substantial. Therefore, it is essential to calculate the test statistic starting from the samples correctly classified in the training set.

\begin{figure}[t]
  \centering
  \includegraphics[width=0.9\columnwidth]{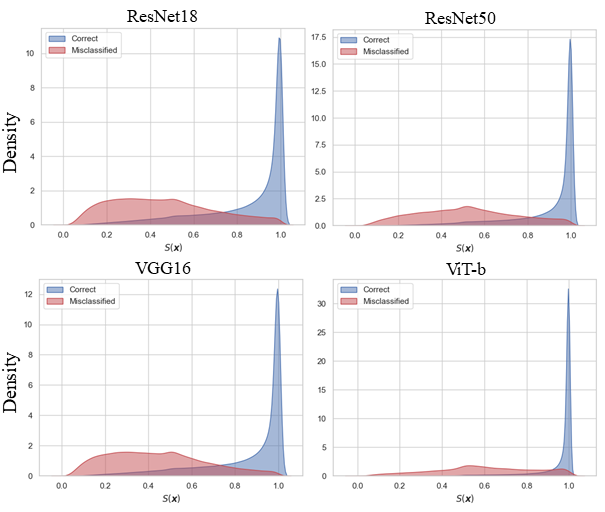}
  \caption{Density plot of ResNet18, ResNet50, VGG16, ViT-b correct and misclassified images. The group correctly classified in the training set and the group misclassified exhibit significant differences. Moreover, according to the probability density curves, these two datasets do not satisfy the normality test, necessitating the choice of the independent samples Mann-Whitney U test.}
  \label{fig:res50}
\end{figure}

Subsequently, based on the classification results of these four preliminary discriminative models on the training set, we plotted the probability density estimation curves for the \(S\) values of samples that were correctly classified and those misclassified within the training set in Fig. \ref{fig:res50}
Here, the red portion represents the misclassified samples, while the blue portion represents the correctly classified samples. The distribution of \(S\) values for the correctly classified samples is primarily concentrated around 0.95, indicating that the preliminary discriminative model is highly confident about these correctly classified samples in the training set, markedly distinguishing them from the misclassified samples.

\end{document}